\documentclass[sigconf]{acmart}

\usepackage[printonlyused]{acronym}
\usepackage{multirow}
\usepackage{ragged2e}
\usepackage{hyperref}
\usepackage{times}  % DO NOT CHANGE THIS
\usepackage{helvet}  % DO NOT CHANGE THIS
\usepackage{courier}  % DO NOT CHANGE THIS
\usepackage{graphicx} % DO NOT CHANGE THIS
\usepackage{natbib}  % DO NOT CHANGE THIS AND DO NOT ADD ANY OPTIONS TO IT
\usepackage{caption} % DO NOT CHANGE THIS AND DO NOT ADD ANY OPTIONS TO IT
\usepackage{siunitx}
\usepackage{booktabs}
\usepackage{makecell}
\usepackage{tikz}
\usepackage{svg}

\usepackage{subcaption} % sub floats
\copyrightyear{2025}
\acmYear{2025}
\setcopyright{cc}
\setcctype{by}
\acmConference[Websci Companion '25]{Proceedings of the 17th ACM Web Science Conference Companion 2025}{May 20--24, 2025}{New Brunswick, NJ, USA}
\acmBooktitle{Proceedings of the 17th ACM Web Science Conference Companion 2025 (Websci Companion '25), May 20--24, 2025, New Brunswick, NJ, USA}
\acmDOI{10.1145/3720554.3733844}
\acmISBN{979-8-4007-1535-8/2025/05}

\begin{document}

%%
%% The "title" command has an optional parameter,
%% allowing the author to define a "short title" to be used in page headers.
\title{Multimodal Misinformation Detection Using Early Fusion of Linguistic, Visual, and Social Features}
%of Linguistic, Visual, and Social Features}

\author{Gautam Kishore Shahi}
%\authornote{Both authors contributed equally to this research.}
\orcid{1234-5678-9012}
\affiliation{%
  \institution{University of Duisburg-Essen}
  \city{Duisburg}
  %\state{Ohio}
  \country{Germany}
}
\email{gautam.shahi@uni-due.de}

%%
%% By default, the full list of authors will be used in the page
%% headers. Often, this list is too long, and will overlap
%% other information printed in the page headers. This command allows
%% the author to define a more concise list
%% of authors' names for this purpose.
%\renewcommand{\shortauthors}{Shahi 2025}

%%
%% By default, the full list of authors will be used on the page
%% headers. Often, this list is too long, and will overlap
%% other information printed in the page headers. This command allows
%% the author to define a more concise list
%% of authors' names for this purpose.
%\renewcommand{\shortauthors}{Trovato et al.}

%%
%% The abstract is a short summary of the work to be presented in the
%% article.

\begin{abstract}
Amid a tidal wave of misinformation flooding social media during elections and crises, extensive research has been conducted on misinformation detection, primarily focusing on text-based or image-based approaches. However, only a few studies have explored multimodal feature combinations, such as integrating text and images for building a classification model to detect misinformation. This study investigates the effectiveness of different multimodal feature combinations, incorporating text, images, and social features using an early fusion approach for the classification model. This study analyzed 1,529 tweets containing both text and images during the COVID-19 pandemic and election periods collected from Twitter (now X). A data enrichment process was applied to extract additional social features, as well as visual features, through techniques such as object detection and optical character recognition (OCR). The results show that combining unsupervised and supervised machine learning models improves classification performance by 15\% compared to unimodal models and by 5\%  compared to bimodal models. Additionally, the study analyzes the propagation patterns of misinformation based on the characteristics of misinformation tweets and the users who disseminate them.
\end{abstract}



%\vspace{-0.4cm}

%%
%% Keywords. The author(s) should pick words that accurately describe
%% the work being presented. Separate the keywords with commas.
%% A "teaser" image appears between the author and affiliation
%% information and the body of the document, and typically spans the
%% page.

%\received{20 February 2007}
%\received[revised]{12 March 2009}
%\received[accepted]{5 June 2009}

%%
%% This command processes the author affiliation and title
%% information and builds the first part of the formatted document.

%\settopmatter{printacmref=false}
%\setcopyright{none}
%\renewcommand\footnotetextcopyrightpermission[1]{}
%\pagestyle{plain}

\settopmatter{printacmref=false}
\maketitle

%\pagestyle{plain}

%\vspace{-0.4cm}

\section{Introduction}
\label{sec:1}

With the growth of digital technology, people are used to getting information online, especially on social media platforms, where users can verify the authenticity of information. Relying on social media for information and news is increasing; however, the rise of mass self-communication, as \citeauthor{castells2010communication} calls it, can bring problems \cite{castells2010communication}, especially during elections or crises where influential actors (such as political candidates) spread information without being factually correct \cite{shahi2024agenda}. Users are open to communicating their ideas and opinions on the platform without any regulations or restrictions, which can lead to spreading misinformation \cite{nathan}. Misinformation influences other users, and they start believing it as true; prior research studied the impact of misinformation and its negative influences on society \cite{hardalov,conroy,drif}. 

%If we look at other studies, there has been a lot of research and work on \ac{VQA}, image captioning, and meme classification in recent years. This is because many real-world problems are multimodal, just as humans experience the world through multimodal senses such as eyes, ears, and tongue. Similarly, data on the Internet and in machines is multimodal, i.e., text, images, video, and sound \cite{afridi}. This proves the importance of applying multimodalities to solve problems, including misinformation. Thus, in this paper, three modalities will be taken into account: text, images, and social features. \\

The spread of misinformation amplifies, especially during crises or elections. \citeauthor{shahi2021exploratory} analyse the spread of misinformation during COVID-19 on Twitter \cite{shahi2021exploratory} and found false tweets spreads faster than true tweets. \citeauthor{yan2025origin} analyzes the role of Artificial Intelligence (AI) in the spread of misinformation during the 2024 US presidential elections. \citeauthor{shahi2025too} analyzes the spread of misinformation during the Russo-Ukrainian conflict on Twitter and the formation of a narrative so that users can believe it \cite{shahi2025too}. 

Prior research solely focused on unimodal data such as text, images, and video individually. However, misinformation appears in different data formats, including multimodal. Social media platforms allow multimodal content, so misinformation can be posted as images or embedded text and images. Misinformation in the form of visuals is more likely than text to stay in our memory, an occurrence known as the "\textit{picture superiority effect}" \cite{childers1984conditions} and the significant impact of visuals in misinformation is undermined \cite{misinformation_intro}. \citeauthor{yilang} provides a theoretical framework for the possible visual attributes that give credibility to the visual features in misinformation \cite{yilang}. \citeauthor{braun} found that the effect of visual features is stronger and more long-lasting than text features. Thus, this research is motivated to focus on multimodal misinformation detection. Hence, in this study, we analyze misinformation posted as images and text during COVID-19 and elections on Twitter.

The present study proposes a fusion-based approach for the detection of multimodal misinformation. The fusion approach uses early fusion and combines different sets of features before feeding to classification models. The present study presents the use of images, social features, and textual features as deciding factors in detecting misinformation on Twitter. In addition, an exploratory analysis is performed to show the characteristics of the users who post misinformation. In this study, the dataset is collected from Twitter using AMUSED framework \cite{shahi2021amused}, which extracts misinformation tweets using fact-checked articles. Then, feature extraction and data enrichment are performed for the classification model. In this study, the following research question is proposed. \\
\textbf{RQ1}: How can we use multimodal classifier models to identify misinformation tweets?\\
%\textbf{RQ2}: What characteristics of fake news and its authors drive its spread of misinformation? 

To answer the first research question, we processed the collected misinformation tweets and extracted text, images, and social features. For misinformation detection, classification models are built to classify misinformation tweets by combining different sets of features, and results analysis is done. Firstly, we run independent unimodal classifier models for each of the modalities and then combine the modalities as we experiment with different multimodal classifier models. The importance of a feature is presented using an exploratory analysis of the results obtained.
%In the experiments, we also try to combine supervised and unsupervised learning in the models to prove if it helps increase the classification model's accuracy. Then,

In order to gain a deeper understanding of the spread of misinformation, different characteristics are used, such as the gender of users posting tweets or whether it is a bot account. Also, user response is measured as retweets, and likes count as diffusion of misinformation.  

The remainder of the paper is organized as follows: Section \ref{sec:2} reviews related work; Section \ref{sec:3} outlines the methodology employed in this study; Section \ref{sec:4} presents the experimental setup, results, and their discussion; and finally, Section \ref{sec:6} concludes the paper and outlines directions for future work.

\section{Related Work}
\label{sec:2}

Text-based misinformation detection is well-researched; it uses multiple textual features that can be extracted from text, including lexical, syntactic, semantic, statistical, and linguistic features.  \citeauthor{alonso} used sentimental analysis for detection of misinformation \cite{alonso}. \citeauthor{hardalov} used linguistic, credibility-related (capitalization, punctuation, pronoun use, and sentiment polarity, with feature selection), and semantic (embeddings and DBpedia entity) features in finding fake news online \cite{hardalov}. 
%Semantic features are used to represent the logical meaning of words in the theoretical way \cite{semantic}. \\
\citeauthor{conroy} provides a typology of different types of truth assessment methods that have emerged from two main categories - linguistic cues with machine learning and network analysis approaches. It is discussed that there is potential in an emerging hybrid method that fuses linguistic cues and machine learning with network-based behavioral data. The linguistic cues or words that people use could be studied for the cases when someone tends to lie; this is being called “predictive deception cues” \cite{conroy}. The results of the study have a high accuracy for classification tasks, however, only within limited domains. Considering we are covering a wider range of domains, this method would not be very beneficial in our paper.

Visual features can also be processed in a few ways, such as forensic features, pixel level, and statistical features. The most common model used for image classification is Convolutional Neural Networks (CNN), as it can be seen used by \citeauthor{kaliyar2020fndnet} \cite{kaliyar2020fndnet}. \citeauthor{juan} examines image forensic features, semantic features, statistical features, and context features for detecting fake news \cite{juan}. It suggests image tampering detection is helpful in identifying if there is manipulation of news. In addition, semantic inconsistencies for logical sense and low image quality may be prevalent and common in fake news. Identifying the image manipulation of images is, however, not so easy to detect in a content-based approach as the changes in the metadata could be subtle. On the other hand, the knowledge-based approach utilizes external resources, which are often untampered images, as a reference to detect image manipulation \cite{juan}. This could be done by searching for the original image source on the internet and retrieving the original metadata of the image. However, in this paper, we apply the content-based approach for the image feature where we only use the given image without external sources from the image database or knowledge of whether the image is original. 
\citeauthor{qi} proposed a framework called MVNN to combine the visual information of the frequency and pixel domains for fake news detection \cite{qi}. The model uses CNN to identify the complex patterns of fake news images in the frequency domain, whereas a multi-branch using CNN and Recurrent neural network (RNN) is used to extract visual cues from distinct semantic layers in the pixel domain. An attention mechanism is applied at the fusion of the feature representations of frequency and pixel domains in order to assign weights to relevant feature representations. The low quality of images and tampered images can be reflected on or represented by the frequency domain \cite{qi}. The effectiveness of the manipulation in visual features is also shown in \cite{imagemanipulation}, where they used error level analysis generated images instead of normal images in order to extract the tampering features. \cite{imagemanipulation} generated good results but stated that text written over images is not considered. These approaches are limited for image-based features without considering text embedded in images, which increases false positive results.

Previous research proposes the use of more than one type of data modality in the detection of misinformation. \citeauthor{wang} proposed an end-to-end framework named EANN, which can deduce event-invariant features and thereby helps to detect fake news in newly incoming events. It is made up of three main components: the multimodal feature extractor, the fake news detector, and the event discriminator. The multimodal feature extractor is used to extract the textual and visual features of tweets and works with the fake news detector to learn the distinct features for the detection of fake news. The event discriminator then distinguishes the common features between the events. The textual and visual modalities are used in this paper.
\citeauthor{raza} proposed a model that is based on a transformer architecture, which has two parts: the encoder part to learn useful features from the fake news data and the decoder part that predicts the future behavior based on past observations \cite{raza}. In this paper, the text and social features modalities are being fed into the model. \citeauthor{jin} proposed a novel RNN with an attention mechanism att-RNN to fuse multimodal features for effective rumor detection \cite{jin}. In this end-to-end network, three modalities are being used that are text, images, and social context. Image features are integrated with the combined features of text and social context that are produced by a Long Short-Term Memory (LSTM) model. The neural attention from the LSTM outputs is leveraged in the fusion with the visual features so as to achieve a robust fused classification. In their paper, the multimodal model has generated promising results \cite{jin}. 

\citeauthor{Antol_2015_ICCV} uses Visual Question Answering (VQA), and the result provides a more granular understanding of the image and more sophisticated reasoning than a system that produces general captions \cite{Antol_2015_ICCV}. \citeauthor{10885260} uses a multimodal fusion approach for the detection of fake news using textual and visual features \cite{10885260}. \citeauthor{shetty2022newscheck} uses OCR extracted from images to classify fake news articles. Overall, OCR has been used individually or in combination with text- and image-based features. However, no prior research has been done in the direction of using multimodal features for the detection of misinformation. 

\section{Methodology}
\label{sec:3}
In this section, the overall pipeline for the detection of misinformation is explained, which includes data collection, data enrichment, and data cleaning and preprocessing, and the classification model is discussed. 

\subsection{Data Collection}
\label{sec:3.1}
Data is collected using AMUSED framework \cite{shahi2021amused}, where misinformation tweets are retrieved from fact-checked articles. Tweets covering misinformation related to elections and COVID-19. The dataset contains 1529 multimodal tweets (combination of image and text) in different languages such as English (67\%), Spanish(16.4\%), French (4.9\%), Portuguese (4.6\%), Hindi (4\%) and others (4\%). Misinformation tweets contain multiple verdicts given by fact-checking organizations. The verdict of misinformation tweets was normalized into four categories (false, true, partially false, and others) following \citeauthor{shahi2021exploratory} \cite{{shahi2021exploratory}}. However, we further merged false and partially false as misinformation and true and others category as non-misinformation. Overall, the dataset is converted for a binary classification. 
Then, a different set of features are extracted from tweets as described below-

%\textbf{Count} & 1026 & 56 & 71 & 251 & 69 & 56 & 1529 \\ 

\subsection{Feature Extraction \& Data Enrichment}
\label{sec:3.2}
In this section, a different set of features and data entrenchment to get more valuable features from data are discussed. Data enrichment or augmentation is the process of enhancing existing information by supplementing missing or incomplete data. Typically, data enrichment is achieved by using external data sources, but that is not always the case \cite{dataenrichment}. Hence, in this work, we perform data enrichment from existing data using different methods, such as OCR. A complete list of features used for model experimentation that are shown in Table \ref{table:socialfeat}.

\subsubsection{Textual features} Textual features are derived from texts of tweets and are converted into word embedding before feeding to the machine learning model. Textual feature includes information such as hashtags and mentions from tweets. 

\subsubsection{Social features} Social feature is defined as the variables obtained from Twitter itself while collecting using the Twitter standard API. As data enrichment, age, gender, and bolometer score were computed as discussed below.

\textbf{Bot Score} Some Twitter handles are created as bots and used for spreading misinformation. Hence, in order to determine if a Twitter handle is a bot, botometer API was used to obtain botometer score \cite{botometer}. The names of the Twitter handle are being fed into the Botometer API provided by Rapid API.\footnote{\url{https://rapidapi.com/OSoMe/api/botometer-pro}} API gives a score in the range of 0 to 5 and feeds into the classification model as the score obtained from Botometer.

%accounts having score more than 2 are considered as bot then bot scores are used as a boolean value, 0 being a human user and 1 a bot. 

\textbf{Gender} For the Twitter handle, gender plays an important role in spreading misinformation \cite{shahi2022mitigating}. In order to classify the gender of account users, a name dictionary that is compiled by \cite{Mejova_2020} using several public name datasets is used. The gender groups that are assigned are male, female, and undetermined. Institutions, groups, and companies are more likely to be assigned to the undetermined group, whereas the other individuals are either male or female.

\textbf{Account Age} Usually, older Twitter accounts are considered stable and not involved in the circulation of misinformation. So, we calculated the account age as the time gap from the date of creation of the account to 2022-09-30. This feature is important as it is safe to say if an account is old and still active, that it has more accountability, and that it is indeed a real account. The oldest account age is 4900 days, while the mean account age is 3750 days. 

\textbf{Popularity of Account} Popularity of account is proposed by \citeauthor{shahi2021exploratory} to measure if accounts are popular on Twitter \cite{shahi2021exploratory}. For a user, popularity is calculated as a ratio of follower counts by following counts, and if it is greater than 1, then the user is popular; otherwise, it is not,  and it is used as a boolean feature.

\begin{table*}[!htbp]
\begin{center}
\caption{Description of different features used in the study}
\begin{tabular}{p{2.5cm}p{8cm}p{2cm}} 
\hline
\textbf{Feature Type} & \textbf{Definition} & \textbf{Representation} \\
\hline
 & Textual Features & \\ \hline
Text  & Text mentioned in the misinformation tweets & Vector \\
Hashtag  & Hashtags mentioned in tweets  & Vector \\
Mention  & User mentioned in tweets  & Vector \\ \hline
  & Social Features & \\ \hline
Created\_at & The UTC datetime when a tweet is posted.  \\
%is quote status & If the post itself is a quote. \\
Retweet Counts & Count of retweet to a tweets  & Numerical \\
Favourite Counts & Count of likes to a tweets & Numerical \\
Retweeted & If the tweet is retweeted & Boolean \\
Followers count & Number of users following a Twitter handle  & Numerical \\
Favorites count & The number of Tweets this user has liked in the account’s lifetime  & Numerical\\
Friends count & The number of users this account is following  & Numerical\\
%Listed count & The number of public lists that this user is a member of. \\
Verified & If a twitter handle is verified by Twitter  & Boolean \\
Statuses count & The number of Tweets (including retweets) issued by the user  & Numerical \\ 
Gender & Gender of users if human else undetermined &  Categorical \\ 
Bot score & Bot score obtained from Botometer & Numerical \\ 
Popularity & Measure popularity of users & Boolean  \\ 
Account Age (days)& Age of accounts in days & Numerical \\ \hline
  & Visual (Image) Feature & \\ \hline
OCR & OCR extract texts from image & Vector \\
Object detection & Object detection from image & Vector \\
\hline
\end{tabular} 
\label{table:socialfeat}
\end{center}
\end{table*}

\begin{table}[h!]
%[pos=ht!]
\centering
\caption{Descriptive analysis of Tweets for both classes}
\label{tab:datasum}
\begin{tabular}{p{4cm}p{1.6cm}p{1.5cm}p{1.5cm}} 
%\toprule
%Dataset: & I & II\\   
\hline
Parameter & False & Other \\
\hline
Number of Tweets  & 1273 & 256   \\  
Unique Account    &  1054 & 229  \\  
Verified Account     & 612 (58\%)  & 125 (55\%) \\   
%Bot  Account  & 22/2(24) &    \\  
Popularity of Account & 939 (78\%) & 205 (80\%)\\
Mean Retweet Count   &  4768 & 4333  \\  
Mean Favourite  Count   & 15706 & 10195  \\  
Mean Followers Count   &  1177680 & 1874661 \\    
Mean Friends Count   &  2935 & 2445  \\  
Mean Status Count & 48008 & 44947 \\
Mean Account Age (days)   & 3801 & 3914   \\
Unique Hashtags & 433 & 94  \\  
Unique mentions &  425 & 84 \\    
Gender(Male/Female/Unknown)&  427/169/458 & 97/27/105\\
\hline
 %\bottomrule
\end{tabular}
\label{table:rumourtheory}
\end{table}

\subsubsection{Visual Feature} A visual feature refers to any characteristic extracted from visual data, such as images or videos, that can used for a classification model. To obtain visual features, we have used OCR, an Object detection from images of misinformation tweets. to retrieve text from the images. 

\textbf{OCR} technology is used to convert virtually any kind of image containing written text (typed, handwritten, or printed) into machine-readable text data \cite{tafti2016ocr}. OCR was implemented using Pytesseract\footnote{\url{https://pypi.org/project/pytesseract/}}, a Python library to extract texts from images. Pytesseract is a wrapper for Google’s Tesseract-OCR Engine.

\textbf{Object Detection} In addition to processing the image, Object mentioned in the images are also identified using Google Object Detection API. The objects detected are used to calculate cosine similarity or correlation between the objects found in the images and the text to see if the captions fit with the images. 
%Object information is however, returned in English only and  translated to various other languages using Google Cloud Translation if needed \cite{cloudvision}.\\
%This could be used to increase the context similarity, such as in \ac{VQA} or \ac{VCR}. The objects detected and the text can be calculated for cosine similarity in order to filter out fake news, in which the images are being misused in a post that is irrelevant, miscaptioned, or false. Oftentimes, the images are being misused to create clickbait for more clicks \cite{fakenewsdef}.\\

%The Google Cloud Vision\footnote{\url{https://cloud.google.com/vision?hl=en}} is used to objects from images. The objects detected could be used to calculate cosine similarity or correlation between the objects found in the images and the text to see if the captions fit with the images. 
%The Google Cloud Vision \ac{API} can detect and extract multiple objects in an image with object localization. "Object localization identifies multiple objects in an image and provides a LocalizedObjectAnnotation for each object in the image. Each LocalizedObjectAnnotation identifies information about the object, the position of the object, and rectangular bounds for the region of the image that contains the object. Object localization identifies both significant and less-prominent objects in an image" \cite{cloudvision}. 
 
\subsection{Data Cleaning \& Preprocessing}
\label{sec:3.3}

The obtained tweets from the above steps are used for language translation into English to have a common language for modeling the textual features. Data is translated using googletrans\footnote{\url{https://pypi.org/project/googletrans/}}, a Python library to use Google Translate for translation of text. Data cleaning is the process of removing noise and unnecessary data. While data cleaning stopwords were removed, URLs were removed. The collected data was highly imbalanced, so four categories were merged into two false (combining false and partially false) and others (combining true and others).

\subsection{Model Architecture}
\label{sec:3.4}

\begin{figure*}[!htbp]
\begin{center}
\includegraphics[width=0.99\linewidth]{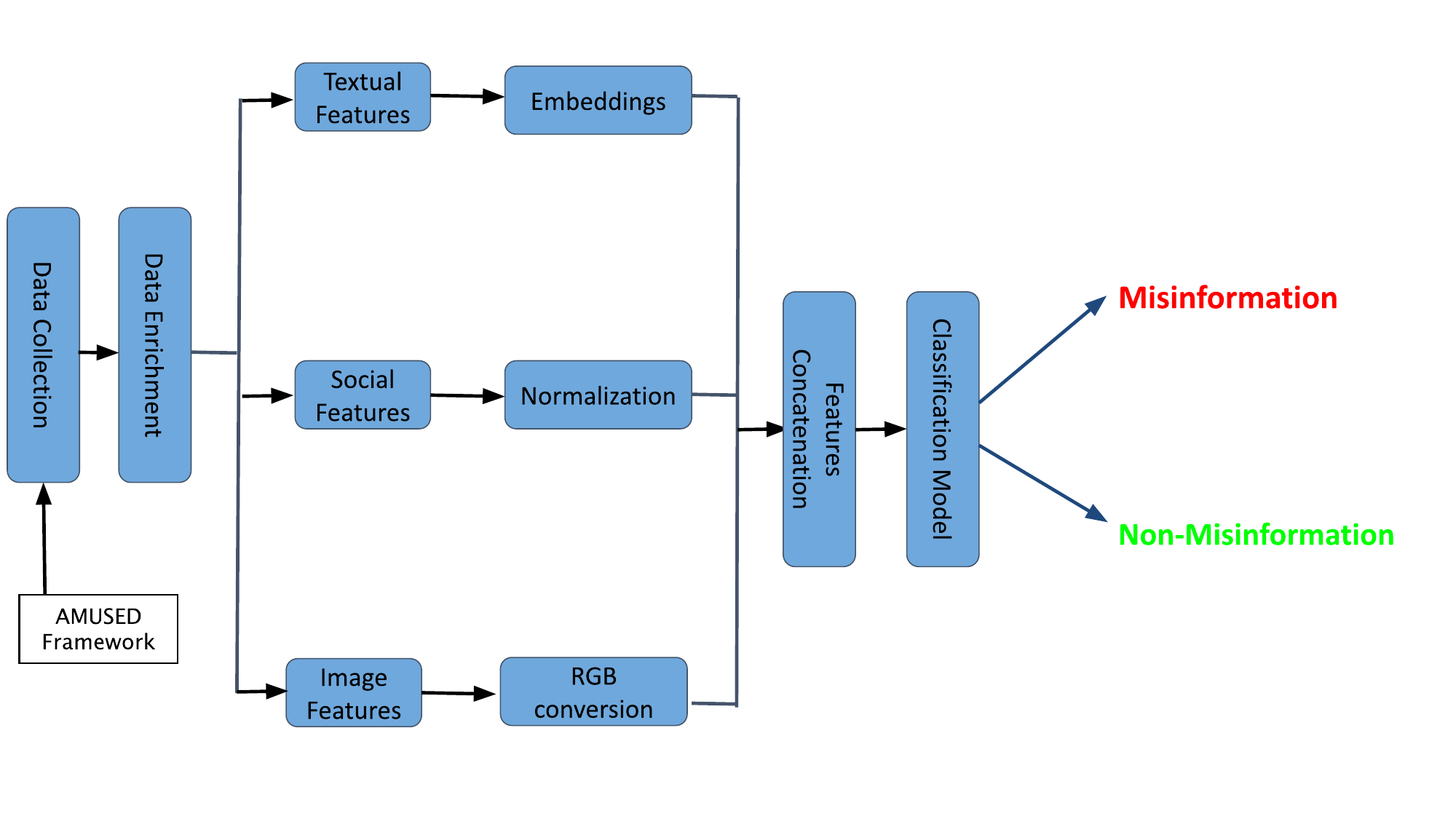} 
\caption{Model Architecture for classification task using all features}
\label{fig:modelarchitecture}
\end{center}
\end{figure*}

The model architecture diagram is shown in Figure \ref{fig:modelarchitecture}. In the figure, we can see that the social features are normalized; this is done to get a range of values that is reasonable in order to prevent data loss. The textual data is converted into vector representation using embeddings. 
%in order for each word to be analyzed. 
The image is also converted to RGB color model for classification models. Overall, an early fusion approach is taken whereby the combinations of modalities are done. As seen in Figure \ref{fig:modelarchitecture}, all the input features are concatenated in a fully connected layer before classification is done, indicating an early fusion approach. If the late fusion approach was to be taken, each of the input features would be independently classified and then the outputs from those classifiers would be concatenated in a later layer to be again classified to get the final output.  

For evaluation of classification results under different combinations of data type, precision, recall, and F1-score are used. Results of different settings are compared using these scores. 

\section{Experiment \& Results}
\label{sec:4}

In this section, we explain the experiment and implementation for the detection of misinformation. The list of features mentioned in section \ref{sec:3.2} 
%are extracted such as hashtags, mentions, bot, popularity. 
Features are represented as mentioned in Table \ref{table:socialfeat} and used for the classification task. We have used state-of-the-art models for comparison of results obtained from the fusion approach. 

\subsection{State-of-the-art Models}
\label{sec:4.1}
In this section, we discuss state of art models used for fake news detection using images. 

%In this section, the state-of-the-art models that are implemented in this paper are going to be introduced and explained in detail. As mentioned in the above subsections, when there are limited computation resources during the training of the models, using pre-trained models would improve the classification results and, at the same time, reduce the computation time and resources needed. Especially in the case of images where a large amount of computational resources are needed to process the data. Under this subsection, the Imagenet and \ac{BERT} pretrained models are discussed as well in detail. 

\textbf{Align Before Fuse (ALBEF)} is a vision language representation learning framework that integrates an image encoder, a text encoder, and a multimodal encoder. ALBEF aligns unimodal image and text representations using an Image-Text Contrastive (ITC) loss before fusing them through cross-modal attention. To enhance multimodal understanding, it employs additional objectives: Image-Text Matching (ITM) to predict whether image-text pairs match and Masked Language Modeling (MLM) to predict masked words using both modalities. To improve learning from noisy web data, ALBEF introduces momentum distillation—a self-training method that leverages pseudo-labels generated by a momentum model (a moving average of the base model). The model is trained using ITC on unimodal encoders and ITM and MLM on the multimodal encoder, with the ITM loss further enhanced through online contrastive hard negative mining.

\textbf{Contrastive Language-Image Pre-training (CLIP)} is a neural network trained on a dataset with a variety of image-text pairs. It can be directed in natural language to predict the most relevant text segment for an image without directly optimizing the image for the task, which is similar to the zero-shot capabilities of the generative pre-trained transformer (GPT)-2 and 3. By jointly training an image and a text encoder, CLIP learns a multimodal embedding space. CLIP maximizes the cosine similarity of the image and text embeddings for the N true pairs in the stack while it minimizes the cosine similarity of the embeddings for the N2-N false pairs. Asymmetric cross-entropy loss is optimized for these similarity values \cite{clip}. 

\textbf{SpotFake} proposed a multimodal framework for fake news detection for image data. The suggested solution uses both the textual and visual features of tweets. They applied the BERT model for training text classification and used the VGG-19 model for image classification in the framework \cite{spotfake}. SpotFake consists of three sub-modules, which are the textual feature extractor, the visual feature extractor, and the multimodal fusion module. The textual feature extractor derives the semantic text features applying a language model, and the visual feature extractor derives the visual features whereby the multimodal fusion module fuses the features obtained from both modalities together to establish a new feature vector.

%\textbf{Imagenet} is a famous project that seeks to provide a large image database for research purposes. Imagenet comprises more than 14 million images and more than 20000 classes. It also provides bounding box annotations for about 1 million images that can be used for object localization tasks. There are many models, such as AlexNet, VGGNet, Inception, ResNet, and Xception. In this work, the VGG-19 model is used as it is found to be the best-performing model for the Imagenet dataset \cite{opencv}. 

\textbf{VGG-19} is a well-known deep convolutional neural network architecture that has demonstrated strong performance in large-scale image recognition tasks. This work utilizes the VGG-19 model, which was originally trained on the ImageNet dataset—a widely used benchmark in computer vision research. ImageNet contains over 14 million annotated images spread across more than 20,000 categories, with around 1 million images also including bounding box annotations for object localization. Several deep learning models have been developed and evaluated using ImageNet, including AlexNet, VGGNet, Inception, ResNet, and Xception. Among these, VGG-19 is selected for this study due to its proven effectiveness and high accuracy on the ImageNet dataset \cite{opencv}.

\textbf{Bidirectional Encoder Representations from Transformers (BERT)} is the pre-trained model that is used for modeling textual features in this study. The trained BERT model is used as a state-of-the-art model by fine-tuning and adding one additional output layer. The state-of-the-art models are used for a broad range of tasks, such as image captioning and question answering, without extensive task-dedicated architecture \cite{DBLP:journals/corr/abs-1810-04805}. Fine-tuning is easy because the self-attention mechanism in the transformer allows BERT to model many downstream tasks, whether they are single text or pairs of text, by exchanging the corresponding inputs and outputs \cite{DBLP:journals/corr/abs-1810-04805}. Hyperparameters in machine learning are values which are used to control the learning process. Parameters are used with a batch size of 32, Adam optimizer, an initial learning rate of 0.1, and loss function as categorical cross-entropy.

\subsection{Results of Classification models}
\label{sec:5}

In this section, we first discuss the results obtained from training different settings, such as unimodal models (using one data type once), bimodal models of input features as a combination of two data types, and lastly, the results of models using all modalities of input features. The results are compared with the types of models experimented by different settings of modalities. 

\subsubsection{Unimodal Results}
\label{sec:5.1}
In Table \ref{table:modelres}, the state of art models are used for classification tasks. For each feature, different models are used, such as for text, BERT, and LSTM models are used; for images, Imagenet and CNN models are used; for social features, CNN and LSTM models are used. 
\begin{table}[h]
\begin{center}
\caption{Classification results of unimodal (one data type at once)}
\begin{tabular}{ cccccc } 
\hline
\textbf{Data} & \textbf{Model} & \textbf{Precision} & \textbf{Recall} & \textbf{F1-score}  \\
\hline
 Text & BERT  & 0.49 & 0.43 & 0.43 \\
 Text & LSTM  & 0.24 & 0.50 & 0.33 \\ \hline
%Image & Imagenet  & 0.24 & 0.49 & 0.32 \\
Image & VGG-19  & 0.24 & 0.49 & 0.32 \\
 Image & CNN  & 0.26 & 0.51 & 0.34 \\ \hline
 Social & CNN  & 0.57 & 0.52 & 0.44 \\
  Social  & LSTM  & 0.24 & 0.49 & 0.32 \\
 \hline
\end{tabular} 
\label{table:modelres}
\end{center}
\end{table}

\noindent Results presented in Table \ref{table:modelres} , overall the CNN using social features classify tweets with highest precision, recall and F1-score, comparing to text and images. For the text modality, BERT  models perform better than LSTM in terms of precision, recall, and F1 score. For image, both Imagenet and CNN perform almost similarly in terms of performance.

\subsubsection{Bimodal}
\label{sec:5.2}

By exploring the results obtained from unimodal classification, the classification model was implemented by combining two input modalities such as image and social, text and social, and image and text. We decided to go with different combinations of models for classification. Table \ref{table:2modelres} presents the results for the models with two input modalities. The combination of BERT and + ALBEF model did the best among all for classification tasks by using text and images, achieving 0.57 for precision, 0.56 for recall, and 0.55 for F1-score. The CNN using image and social features was the least performant, with only 0.49 for accuracy, 0.48 for precision, 0.40 for recall, and 0.48 for F1-score. All of the other models show, in general, an improvement from the unimodal models.  

\begin{table*}[!htbp]
\begin{center}
\caption{Classification results of unimodal (combination two data types at once)}
\begin{tabular}{ cccccc } 
\hline
\textbf{Modalities} & \textbf{Model} & \textbf{Accuracy} & \textbf{Precision} & \textbf{Recall} & \textbf{F1-score}  \\
\hline
 Image+Social & CNN & 0.49 & 0.48 & 0.49 & 0.48 \\
 Text+Social & BERT+CNN & 0.55 & 0.61 & 0.55 & 0.5 \\
 Text+Image & BERT+CNN & 0.54 & 0.54 & 0.54 & 0.54 \\
 Text+Image & BERT+ALBEF & 0.56 & 0.56 & 0.55 & 0.54 \\
  %Text+Image & BERT+ALBEF & 0.56 & 0.57 & 0.56 & 0.55 \\
 Text+Image & BERT+CLIP & 0.52 & 0.52 & 0.52 & 0.52 \\ \hline
\end{tabular} 
\label{table:2modelres}
\end{center}
\end{table*}

\subsubsection{Three Modalities}
\label{sec:5.3}
Finally, various combinations of all three data modalities—text, images, and social features—were explored for the classification task by using different combination of models. The performance of four combination of models is summarized in Table \ref{table:3modelres}. Interestingly, models that incorporated all three modalities did not show significant improvement over those using only two, with the exception of the CLIP+CNN model. 

\begin{figure}[h!]
\begin{center}
\includegraphics[width=0.99\linewidth]{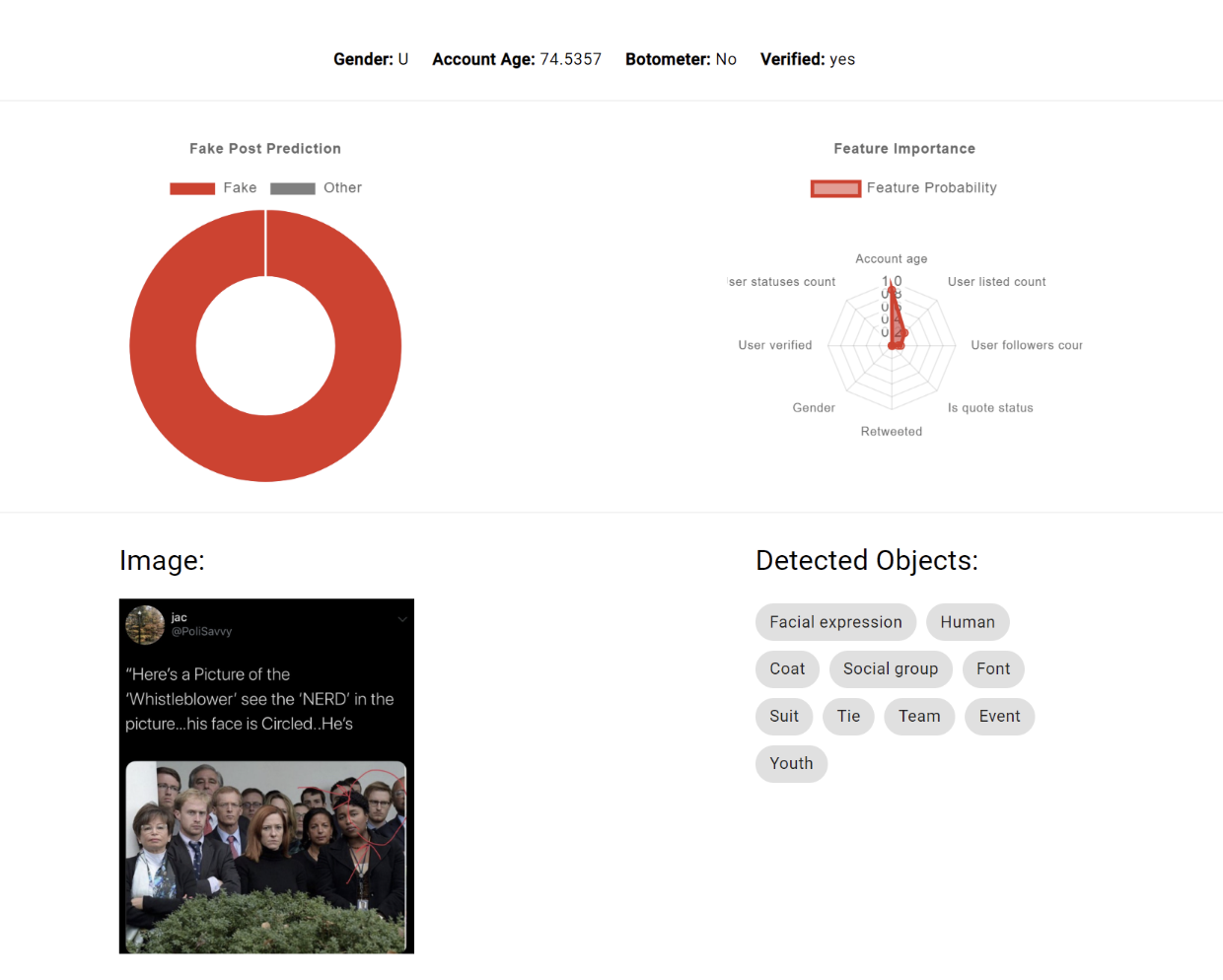} 
\caption{Detection of misinformation tweet}
\label{fig:result}
\end{center}
\end{figure}

Classification models combining three data types using CLIP and CNN achieve the highest overall scores in terms of accuracy, precision, recall, and F1-score. It is important to highlight that CLIP operates as an autoencoder employing unsupervised learning to assess image authenticity, while CNN is a supervised learning model applied to social features and textual data. The synergy between unsupervised and supervised models in the CLIP+CNN configuration demonstrates strong potential for effective misinformation classification. The best-performing models outperform bimodal by 5\% and 15\% for unimodal classification models. This leads us to the conclusion that the modalities used as the input features were all in some way helpful to the classification models. A detection of misinformation tweets and explaining all features is presented in Figure \ref{fig:result}.

\begin{table*}[!htbp]
\begin{center}
\caption{The three modalities' model results}
\begin{tabular}{ cccccc } 
\hline
\textbf{Modalities} & \textbf{Model} & \textbf{Accuracy} & \textbf{Precision} & \textbf{Recall} & \textbf{F1-score}  \\
\hline
 Text+Image+Social & VGG-19+BERT+CNN & 0.49 & 0.24 & 0.49 & 0.32 \\
 Text+Image+Social & CNN+BERT+CNN & 0.47 & 0.47 & 0.47 & 0.47 \\
 Text+Image+Social & ALBEF+CNN & 0.56 & 0.57 & 0.56 & 0.54 \\
 Text+Image+Social & CLIP+CNN & 0.59 & 0.60 & 0.59 & 0.59 \\ \hline
\end{tabular} 
\label{table:3modelres}
\end{center}
\end{table*}

In addition, transfer learning or fine-tuning the models with a larger pre-trained dataset did not necessarily generate better results. For BERT in the text-based unimodal model, the model results are stable; however, they are not far better than the results of the text-based LSTM model. For the image-based unimodal model, the pretrained model imagenet even scored a lower model result compared to a CNN model. However, it is noted that the combination of the unsupervised and supervised machine learning models did increase the overall performance of the model. The usage of the ALBEF and CLIP models as unsupervised learning models shown in Table \ref{table:3modelres} proved that when the ALBEF and CLIP models are fused with CNN, they produce far better results than other models, which just use supervised learning. 

\subsection{Propagation of Misinformation}
\label{sec:5.4}

Misinformation tweets and users who posted them were analyzed for deeper insights into misinformation propagation. Table \ref{table:rumourtheory} shows the descriptive analysis of collected tweets. Users are analyzed based on gender, bots, verified accounts, and popularity. The spread of misinformation is analyzed based on retweets and likes as a measure of the spread of information diffusion \cite{stieglitz2013emotions}. 

Regarding users who posted misinformation, there are no overall differences on different parameters mentioned in Table \ref{table:rumourtheory} except users posting false tweets have fewer followers and get more likes counts. However, after investigating in deeper each characteristic, verified accounts have a huge follower base, and tweets posted by verified accounts get more than double the retweets and almost three times as many likes for misinformation, which helps to spread faster in the network a similar trend was observed by \citeauthor{shahi2021exploratory} \cite{shahi2021exploratory}. In terms of gender, false news posted by male users gets more than double in terms of likes and retweets for tweets. In the case of other category, the pattern is reversed. Hence, male misinformation tweets spread faster than other kinds of tweets. In the case of popularity, false tweets get more likes and retweets than other tweets, irrespective of users' popularity. 

To summarise, humans tend to spread false information and get more attention from other users based on different user characteristics. However, even if accounts are verified, we need to be aware because they can also spread misinformation. As mentioned before, we have witnessed the effects of misinformation on social media that could influence even wars \cite{shahi2025too} and elections \cite{shahi2024agenda}. So, a user should be careful before believing or circulating tweets that can lead to misinformation diffusion.

\section{Conclusion \& Future Work}
\label{sec:6}

The present study explores the use of multimodal features for the detection of misinformation and tests for misinformation tweets collected from COVID-19 and elections. Combining different features improves the classification performance by 15\% for unimodal and 5\% for bimodal. Classification models are tested using small datasets, which can be improved using large datasets. In addition, the data obtained from fact-checking organizations mainly contributes to false and partially false categories, which makes the data unbalanced. Since this limitation is real and common, it is often mentioned that it would be worthwhile to research unsupervised training and learning with unbalanced data. In terms of feature analysis, different features are extracted and useful for the classification model. However, the bot score did not give any promising score, so it was hard to say if there is any bot accounts were used in the study. 
%The combination of supervised and unsupervised machine learning models in the classifier helped greatly because data annotation in the misinformation is costly and time-consuming. In addition, the fake news domain is too large, making it difficult for supervised machine learning models to learn a specific type of pattern across a wide range of domains. The dataset used in our paper included domains from politics, war, and pandemics. Models trained on datasets in specific domains only might produce better results. 
As future work, this paper only considers images, text, and social features as model inputs. Videos are not considered. The tweets with videos might still be able to be predicted by the fake news detection tool; however, the uncertainty of which frame of the video might be processed as the image may well cause the results to be inaccurate. Another interesting extension of this work is to implement and research the misinformation detection model on other social media platforms for wider use and availability. Furthermore, this paper only explores the early fusion of modalities in the classification models. Advanced fusion approaches can be explored for the classification of misinformation. 

\bibliographystyle{ACM-Reference-Format}
\bibliography{multimodal}

\end{document}